\documentclass{article}

\PassOptionsToPackage{numbers, compress}{natbib}
\usepackage{natbib}  

\usepackage{enumitem}
\usepackage{graphicx}

\usepackage[utf8]{inputenc} 
\usepackage[T1]{fontenc}    
\usepackage{hyperref}       
\usepackage{url}            
\usepackage{booktabs}       
\usepackage{amsfonts}       
\usepackage{amsmath}
\usepackage{amssymb}
\usepackage{nicefrac}       
\usepackage{microtype}      
\usepackage[table]{xcolor}  
\usepackage{enumitem}
\usepackage{pifont}
\usepackage{graphicx}
\newcommand{\cmark}{\ding{51}}%
\newcommand{\xmark}{\ding{55}}%
\usepackage{array}

\usepackage{changepage,threeparttable} 
\usepackage{multirow}
\usepackage{caption}
\usepackage{subcaption}
\usepackage{lipsum}
\usepackage{graphicx}
\usepackage{flushend}
\usepackage{xspace}
\usepackage{wrapfig}
\usepackage{algorithm}
\definecolor{newblue}{HTML}{1770b3}
\definecolor{newgreen}{HTML}{00CC66}
\definecolor{newpurple}{HTML}{A020F0}
\definecolor{newred}{HTML}{CC0000}
\definecolor{newpink}{HTML}{FFB6C1}
\definecolor{blue_new}{rgb}{0.06, 0.75, 0.99}
\definecolor{LightCyan}{rgb}{0.88,1,1}
\newcommand{\greenup}{\textcolor{newgreen}{\boldsymbol{\uparrow}}}
\newcommand{\redown}{\textcolor{red}{\boldsymbol{\downarrow}}}
\hypersetup{
    colorlinks,
    linkcolor={red},
    citecolor={newblue},
    urlcolor={blue!80!black}
}

\usepackage[preprint]{neurips_2024}

\title{Rethinking Visual Prompting for Multimodal Large Language Models with External Knowledge}

\author{
  Yuanze Lin$^\clubsuit$\thanks{Work done during an internship at Microsoft Redmond.} \quad\quad Yunsheng Li$^\spadesuit$ \quad\quad \textbf{Dongdong Chen}$^\spadesuit$ \quad\\ \textbf{Weijian Xu}$^\spadesuit$ \quad\quad \textbf{Ronald Clark}$^\clubsuit$ \quad\quad \textbf{Philip Torr}$^\clubsuit$ \quad\quad \textbf{Lu Yuan}$^\spadesuit$ \\
  $^\clubsuit$ University of Oxford \quad\quad $^\spadesuit$Microsoft\\
}

\begin{document}

\maketitle

\begin{abstract}
In recent years, multimodal large language models (MLLMs) have made significant strides by training on vast high-quality image-text datasets, enabling them to generally understand images well. However, the inherent difficulty in explicitly conveying fine-grained or spatially dense information in text, such as masks, poses a challenge for MLLMs, limiting their ability to answer questions requiring an understanding of detailed or localized visual elements. Drawing inspiration from the Retrieval-Augmented Generation (RAG) concept, this paper proposes a new visual prompt approach to integrate fine-grained external knowledge, gleaned from specialized vision models (\textit{e.g.}, instance segmentation/OCR models), into MLLMs. This is a promising yet underexplored direction for enhancing MLLMs' performance. Our approach diverges from concurrent works, which transform external knowledge into additional text prompts, necessitating the model to indirectly learn the correspondence between visual content and text coordinates. Instead, we propose embedding fine-grained knowledge information directly into a spatial embedding map as a visual prompt. This design can be effortlessly incorporated into various MLLMs, such as LLaVA and Mipha, considerably improving their visual understanding performance. Through rigorous experiments, we demonstrate that our method can enhance MLLM performance across nine benchmarks, amplifying their fine-grained context-aware capabilities.

\end{abstract}
\section{Introduction}

The advancement of large language models (LLMs)~\cite{touvron2023llama,chatgpt,gpt4,Google_Bard} has revolutionized how machines process and generate human-like text, demonstrating remarkable abilities in reasoning, translation, and contextual understanding. The integration of language and vision into unified models, such as GPT-4V~\cite{GPT4V_System_Card}, represents a significant leap forward in enabling machines to understand and interact with the world in a manner akin to human cognition. As these models continue to evolve, they promise to further blur the lines between human and machine cognition, opening new frontiers in AI research and application~\cite{lin2024text,rombach2022high, poole2022dreamfusion,lin2024dreampolisher,team2023gemini,lin2023smaug}.

\begin{figure*}[t]
\centering
\includegraphics[width=1\linewidth]{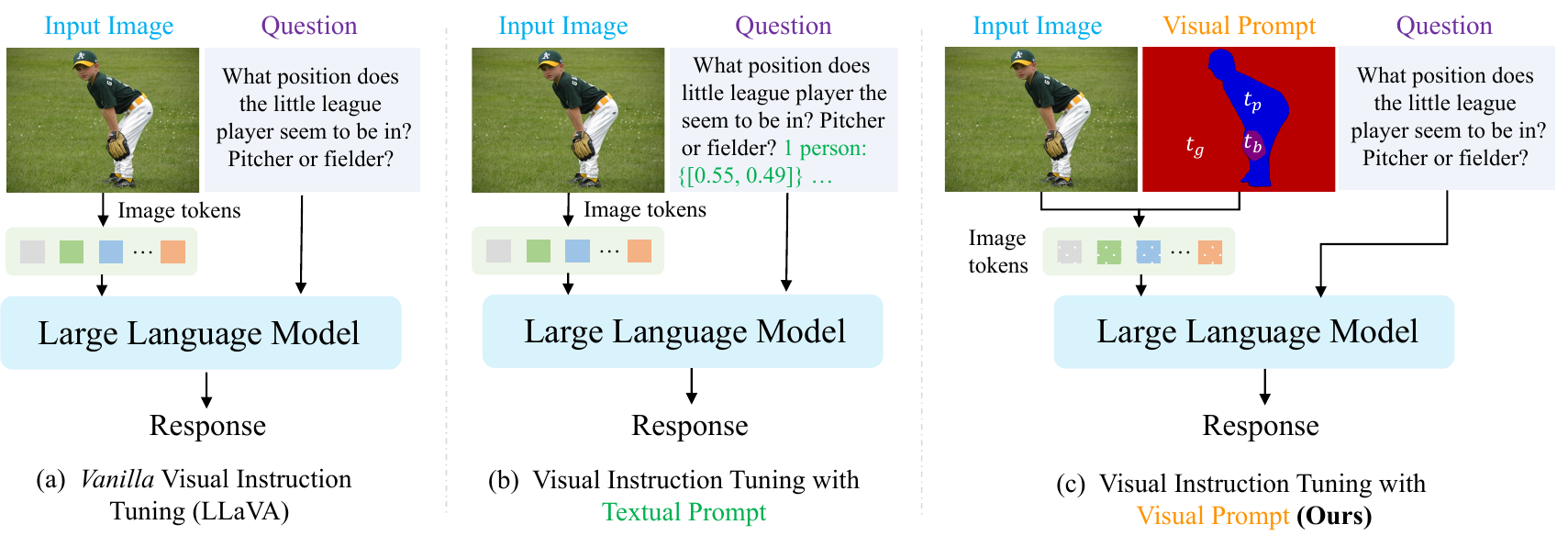}
\vspace{-4mm}
    \caption{\textbf{Different training paradigms.} (a) means the original visual instruction tuning of LLaVA~\cite{liu2023improved}. (b) denotes visual instruction tuning with external textual prompts ~\cite{jiao2024enhancing} (\textit{e.g.}, \textcolor{newgreen}{1 person} and the center coordinates of its bounding box: \textcolor{newgreen}{[0.55,0.49]}), note that we neglect the template prefix of textual prompts for visualization. (c) is the proposed  auxiliary visual prompt, which is a feature map composed with different object regions. For each pixel, it is filled out with the textual embedding of the corresponding \textit{categories} or \textit{OCR text} ($t_{\mathrm{g}}$, $t_{\mathrm{p}}$ and $t_{\mathrm{b}}$ in the example visual prompt mean the textual embeddings of \textcolor{newred}{\textit{grass}}, \textcolor{blue}{\textit{person}} and \textcolor{newpurple}{\textit{baseball glove}}).}
\label{fig:figure1}
\vspace{-4mm}
\end{figure*}

Despite their remarkable capabilities, most of the MLLMs (shown in Figure \ref{fig:figure1} (a)) trained with image-text pairs still often struggle in  fine-grained multimodal comprehension capacities, \textit{e.g.}, correctly count objects or output precise location of one specific object. This is partially because of the lack of high-quality data with exceptionally fine-grained text description. More importantly, text itself has the inherent difficulty in accurately conveying highly fine-grained or spatially dense information. As a result, current MLLMs often fail to accurately interpret pixel-level visual content of localized regions within an image, which in return harms the overall comprehension capacity for the image and thereby causes the notorious ``hallucination'' problem~\cite{li2024multimodal}.

To tackle this challenge, one line of work ~\cite{chen2023minigptv2,zhao2023chatspot,cai2023making} explicitly integrates region coordinates information into the text prompt  and trains on specialized region-level chatting data. However, this still demands that the model implicitly learns to understand coordinates and establish connections with visual content, thereby increasing the learning complexity. Another line of work ~\cite{rasheed2023glamm,zhang2023gpt4roi,lin2022revive} proposes incorporating Region of Interest (ROI) features directly into model learning, necessitating bespoke model architectures. In contrast to these approaches, rather than starting from scratch to learn region information, this paper explores leveraging finely-grained recognition outcomes directly obtainable from existing vision models as external knowledge for MLLMs, inspired by the RAG concept. Concurrent with our work, one recent approach  ~\cite{jiao2024enhancing}  introduces external knowledge, such as regional coordinates from object detection and Optical Character Recognition (OCR) technologies, into MLLMs (shown in Figure \ref{fig:figure1} (b)), helping understand localized multimodal content. However, this method still integrates external knowledge through the text prompt, mandating implicit learning of content-to-coordinate correspondence by the model. Furthermore, it lacks support for more nuanced external knowledge, such as instance masks.

In this paper, we propose a new visual prompt paradigm to insert external knowledge, \textit{e.g.}, localized information, into MLLMs addressing the challenge of fine-grained multimodal content correspondence. As illustrated in Figure \ref{fig:figure1} (c), the core idea is, rather than treating local context information as a part of text prompts, we embed them directly within the visual prompts. Specifically, we start by leveraging panoptic segmentation~\cite{zhang2023simple} and OCR detection~\cite{du2021pp} models, and a pre-trained text encoder to generate pixel-wise text embeddings, which are served as the local context information for MLLMs. Subsequently, we extend the original visual prompts by adding the newly generated context information in a spatial-wise manner. This integrated prompt is then assimilated into MLLMs, improving fine-grained visual content comprehension. Consequently, our approach is capable of enabling MLLMs to discern contexts in the pixel-level space and improve their performance.

With the proposed visual prompt paradigm, we train a bunch of MLLMs on the LLaVA-1.5 datasets~\cite{liu2023improved}. The experimental results show that, even with 3 billion parameters, our method improves upon the leading open-source MLLMs such as LLaVA-1.5~\cite{liu2024visual,liu2023improved} and Qwen-VL~\cite{bai2023qwen}, without needing additional training data. Remarkably, our models showcase superior performance across a wide array of benchmarks when compared to the 7-billion MLLM variants, including LLaVA-1.5, Qwen-VL, and InstructBLIP~\cite{dai2024instructblip}, and in some instances, even outperform their 13-billion MLLM counterparts. Our experimental results confirm the significance of integrating our proposed prompt approach with MLLMs to enhance cognitive capabilities.

The contributions can be summarized as follows:
\begin{itemize}
\item We systematically investigate integrating localized information into MLLMs. Empirical findings suggest that our proposed visual prompt significantly outperforms the previous prompt paradigm relying solely on textual prompts containing coordinates.

\item We propose to integrate contextual embeddings within local contours (\textit{e.g.}, object masks) as the visual prompt, which facilitates the establishment of correlations between image pixels and contexts, thereby enhancing the fine-grained cognitive capabilities of various MLLMs across a spectrum of benchmarks.

\item Based on our proposed approach, our model with 3B parameters surpasses both existing 7B and 13B models across diverse benchmarks, all without the need for extra training data. 
\end{itemize}

\begin{figure*}[t]
\centering
\includegraphics[width=1\linewidth]{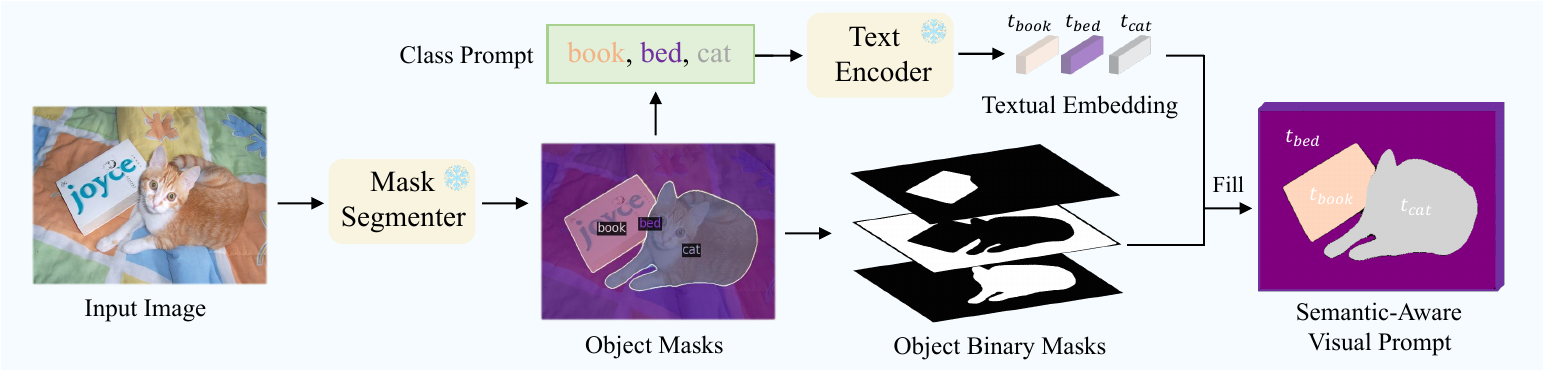}
\vspace{-4mm}
    \caption{\textbf{Auxiliary visual prompt generation}. It firstly generates the panoptic segmentation masks~\cite{zhang2023simple} for the input image, there's a class category for each mask region, then we can obtain the textual embeddings (\textit{e.g.}, $t_{\mathrm{book}}$, $t_{\mathrm{bed}}$ and $t_{\mathrm{cat}}$) through a pre-trained text encoder for all the classes (\textit{e.g.}, \textcolor{newpink}{book}, \textcolor{newpurple}{bed}, \textcolor{gray}{cat}). Finally, the auxiliary visual prompt can be generated by concatenating these textual embeddings within the corresponding mask regions together. Note that we can also adopt the OCR model~\cite{du2021pp} to obtain the texts and the regions, we don't display it here for clearer explanation.} 
\label{fig:figure2}
\end{figure*}

\begin{figure*}[t]
\centering
\includegraphics[width=0.85\linewidth]{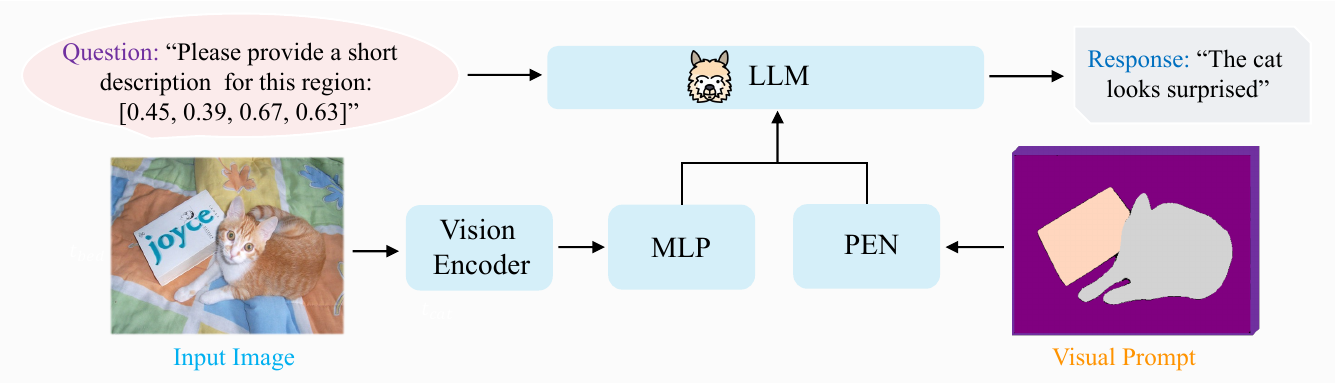}
    \caption{\textbf{The illustration of visual instruction tuning with the generated visual prompt.} Our proposed visual prompt can be easily combined with existing multimodal large language models (\textit{e.g.}, LLaVA~\cite{liu2023improved}), note that \textit{PEN} means prompt embedding network.}
\label{fig:figure3}
\end{figure*}
\section{Related Work}
\paragraph{Large Language Models.} The initial potential of large language models (LLMs) was showcased by foundational works like BERT~\cite{devlin2018bert} and GPT~\cite{radford2018gpt}. They sparked a wave of scaling efforts, leading to a range of influential projects, such as T5~\cite{raffel2020t5}, GPT-3~\cite{brown2020gpt3}, Flan-T5~\cite{chung2022flant5}, and PaLM~\cite{chowdhery2022palm}. As the volume of training data expanded and the dimensions of model parameters grew, these scaling endeavors led to the creation of ChatGPT~\cite{openai2022chatgpt,ouyang2022instructgpt}. Models like LLaMA~\cite{touvron2023llama} and GPT-4~\cite{gpt4} have been trained on extensive corpora and demonstrated remarkable capabilities in diverse cognitive tasks. Additionally, lightweight LLMs with fewer than 3B parameters, \textit{i.e.}, Phi~\cite{abdin2024phi,microsoft2024phi2} and StableLM-2~\cite{stablelm2023} have shown performance comparable to larger models~\cite{vicuna}. In our work, we adopt Phi-2~\cite{microsoft2024phi2}  and Vicuna-7B~\cite{vicuna} as our language backbone.

\paragraph{Multimodal Large Language Models. } Influenced by the success of instruction tuning from LLM, LLaVA~\cite{liu2024visual} and MiniGPT-4~\cite{minigpt4} have adopted visual instruction tuning to improve LLMs' interaction with visual data, yielding impressive outcomes. Kosmos-2~\cite{peng2023kosmos} and Shikra~\cite{chen2023shikra} have advanced MLLMs by enhancing visual comprehension capabilities. while works like LLaVA-Phi~\cite{zhu2024llava}, MobileVLM~\cite{chu2024mobilevlm} and Bunny~\cite{he2024efficient} mainly focus on optimizing training recipes and architecture design for lightweight MLLMs. To solve the challenge of understanding fine-grained information in images, existing approaches propose to learn coordinate representations~\cite{chen2023minigptv2,chen2023shikra,zhao2023chatspot} and Region of Interest (ROI) features ~\cite{peng2023kosmos,zhang2023gpt4roi}, which use inflexible visual referral formats or necessitate the collection of region-level training data. On the contrary, we focus on utilizing external knowledge to improve the fine-grained vision-language alignment for MLLMs without collecting extra chatting data.

\paragraph{Prompting Multimodal Large Language Models.} Inspired by the ability of GPT-4V~\cite{GPT4V_System_Card} to process diverse inputs, ViP-LLaVA~\cite{cai2023making} collects a visual prompt instruction dataset containing various visual prompts, \textit{e.g.}, scribbles and arrows, for MLLM fine-tuning. Contemporary to our work, ~\cite{jiao2024enhancing} has offered advanced insights in prompting MLLMs through external knowledge, which introduces bounding box and OCR coordinates into text prompt, however, it's still challenging to interpret the pixel-level contexts. In this paper, we investigate how to efficiently utilize external knowledge to enhance multimodal fine-grained alignment of MLLMs and introduce a novel visual prompt paradigm incorporating pixel-level contextual information.
\section{Proposed Method}
In this section, we propose a new visual prompt paradigm that integrates local external information to enhance the capability of MLLMs. In section~\ref{instruction}, we outline the design of the auxiliary visual prompt that contains local contextual information. Using the auxiliary visual prompt, in section~\ref{infusion}, we further embed it into MLLMs by merging it with the original visual tokens. Finally, we briefly introduce the details of training in section~\ref{training}.

\subsection{Auxiliary Visual Prompt with External Knowledge}
\label{instruction}
In this section, we propose a method to generate local contextual external knowledge to assist MLLMs. In contrast to \cite{jiao2024enhancing}, which focuses solely on object detection and OCR information and integrates them as part of the text prompt, we enhance the granularity of local external knowledge by leveraging a panoptic segmentation model. Additionally, we continue to utilize an OCR model but transform both types of external knowledge into pixel-wise embeddings. Further details are provided below.

As shown in Figure~\ref{fig:figure2}, given the input image $I \in \mathbb{R}^{3\times H \times W}$, we can obtain the fine-grained external knowledge by an off-the-shelf panoptic segmentation model~\cite{zhang2023simple} and an OCR model~\cite{du2021pp}. The generation of the external knowledge can be expressed as:
\begin{equation}
    \{M_{j}, C_{j}\}^{N_{s}}_{j=1} = f_{\mathrm{seg}}(I), \hspace{4mm}
    \{{B}_{j}, {T}_{j}\}^{N_{o}}_{j=1} = f_{\mathrm{ocr}}(I),
\end{equation}
where $f_{\mathrm{seg}}(\cdot)$ and $f_{\mathrm{ocr}}(\cdot)$ mean panoptic segmentation and optical character recognition (OCR) models, $N_{\mathrm{s}}$ and $N_{\mathrm{o}}$ are the numbers of detected mask regions and OCR bounding boxes. $\{M_{j}, C_{j}\}^{N_{\mathrm{s}}}_{j=1}$ is the set of mask regions and the corresponding classes, and $\{B_{j}, T_{j}\}^{N_{\mathrm{o}}}_{j=1}$ represents the set of detected OCR bounding boxes and texts.

With the detected classes $\{C_{j}\}^{N_{\mathrm{s}}}_{j=1}$ and OCR texts $\{T_{j}\}^{N_{\mathrm{o}}}_{j=1}$, a pre-trained text encoder ($f_{\mathrm{text}}(\cdot)$) is leveraged to generate the texture embeddings as:

\begin{equation}
  \label{equation2}
  \begin{aligned}
    \mathcal{T_{\mathrm{s}}} = \{t_{1}, \ldots, t_{N_{\mathrm{s}}}\} = \{f_{\mathrm{text}}(C_{1}), \ldots, f_{\mathrm{text}}(C_{N_{\mathrm{s}}})\}, \\        
    \mathcal{T_{\mathrm{o}}} = \{\hat{t}_{1}, \ldots, \hat{t}_{N_{\mathrm{o}}}\}= \{(f_{\mathrm{text}}(T_{1}), \ldots, f_{\mathrm{text}}(T_{N_{\mathrm{o}}})\},
  \end{aligned}
\end{equation}

where $t_{i} \in \mathbb{R}^{1 \times d} (1 \leqslant i \leqslant N_{\mathrm{s}})$ and $\hat{t}_{i} \in \mathbb{R}^{1 \times d} (1 \leqslant i \leqslant N_{\mathrm{o}})$ denote the $i_{\mathrm{th}}$ textual embedding vector of the classes for the detected mask region and OCR texts respectively, while $d$ is the embedding dimension.

In order to generate a pixel-wise visual prompt for the external knowledge instead of a pure text description for the regions with coordinates and category names, the auxiliary visual prompt is initialized as a zero tensor $\mathcal{P} \in \mathbb{R}^{H \times W \times d}$ and then filled with the newly generated texture embeddings for the external knowledge as:

\begin{equation}
\begin{aligned}
\mathcal{P}_{j, k} = 
\begin{cases}
t_u & \text{if } (j, k) \in M_u  \\
\mathcal{P}_{j, k} & \text{otherwise}
\end{cases}
\quad \forall u \in \{1, \ldots, N_{\mathrm{s}}\},
\\
\mathcal{P}_{j, k} = \mathcal{P}_{j, k} +
\begin{cases}
\hat{t}_v & \text{if } (j, k) \in B_v  \\
0 & \text{otherwise}
\end{cases}
\quad \forall v \in \{1, \ldots, N_{\mathrm{o}}\}.
\end{aligned}
\end{equation}

Note, for some regions, if the confidence of the class prediction given by the segmentation model is low or the OCR model fails to detect any text, we leave the region area with zero values. For the regions that are occupied by both model, we simply add the text embeddings directly. We leave the investigation of more refined fusion techniques to future research.

With the auxiliary visual prompt containing pixel-level local contextual information from panoptic segmentation and OCR models, MLLMs can effectively capture finer-grained features. The next challenge is to establish a clearer connection between the newly generated external knowledge and the original image feature. This will help alleviate the model's difficulties in learning their relationship effectively.

\subsection{Visual Prompt Infusion}
\label{infusion}
In this section, we introduce the visual prompt infusion that incorporates the proposed auxiliary visual prompts into the MLLMs. Previous methods~\cite{jiao2024enhancing} choose to append the external knowledge (embeddings for object category and its coordinates) to the text prompts, which requires the model to learn the correspondence of visual content within the specified coordinates encoded in the external knowledge and, as a result, increasing the difficulties of the learning process of the model. To address this challenge, we propose to merge the auxiliary visual prompt directly with the image features in a pixel-wise manner.

Specifically, as shown in Figure~\ref{fig:figure3}, the image tokens are first generated via an image encoder $f_{\mathrm{img}}(\cdot)$ and an MLP projector ($f_{\mathrm{MLP}}(\cdot)$):
\begin{equation}
\label{equation4}
    \mathcal{F}_{\mathrm{v}} = f_{\mathrm{MLP}}(f_{\mathrm{img}}(I)),
\end{equation}
where $\mathcal{F}_{\mathrm{v}} \in \mathbb{R}^{N_{\mathrm{v}} \times d_{\mathrm{v}}}$, $N_{\mathrm{v}}$ and $d_{\mathrm{v}}$ represent the number of image tokens and the embedding dimension. Then, the auxiliary visual prompt is further processed by a prompt embedding network (PEN) as

\begin{equation}
        \mathcal{F}_{\mathrm{p}} = f_{\text{PEN}}(\mathcal{P}).
\end{equation} 
For the prompt embedding network, we employ three convolutional layers, with an activation layer (ReLU) inserted between each pair of them. This network primarily serves to align the feature space and spatial size between the image tokens and the auxiliary visual prompts.

When combining the image tokens and the processed auxiliary visual prompt, we mainly consider two options, both of which operate pixel-wise. 
\textbf{(1) feature fusion:} $\hat{\mathcal{F}_{\mathrm{v}}} = f(\mathrm{Concat}(\mathcal{F}_{\mathrm{v}}, \mathcal{F}_{\mathrm{p}}))$, where $f$ is a linear layer that maps the embedding $\mathbb{R}^{N_{v} \times d_{2v}} \rightarrow \mathbb{R}^{N_{v} \times d_{v}}$ to maintain the total number of image tokens unchanged; \textbf{(2) feature addition}, $\hat{\mathcal{F}}_{\mathrm{v}} = \mathcal{F}_{\mathrm{v}} + \mathcal{F}_{\mathrm{p}}$, which sums the two types of features directly. 

The advantages of the pixel-wise infusion for both options facilitate the model's comprehension of the correspondence between external knowledge and original visual features. This explicit guidance enables the model to easily understand the pixel categories as well as the potential OCR text description it conveys. Consequently, it aids the model in disambiguating complex scenes, accentuating salient features, and distinguishing finer objects.

\subsection{Training}
\label{training}

Training MLLMs involves predicting responses based on multimodal inputs using an autoregressive approach. The objective is to maximize the probability of generating tokens that match the ground-truth answer $Y_{\mathrm{a}}$. With the new visual embedding $\hat{\mathcal{F}}_{\mathrm{v}}$, this can be mathematically expressed as follows:

\begin{equation}
P(Y_{\mathrm{a}} | \hat{\mathcal{F}_{\mathrm{v}}}, \mathcal{F}_{\mathrm{t}}) = \prod_{i=1}^{L} P_{\boldsymbol{\theta}} (y_{i} | \hat{\mathcal{F}_{\mathrm{v}}}, \mathcal{F}_{\mathrm{t}}, Y_{\mathrm{a},<i}).
\end{equation}
Here, $L$ represents the sequence length of the ground truth answer $Y_{\mathrm{a}}$, $\boldsymbol{\theta}$ means the trainable parameters. $Y_{\mathrm{a},<i}$ represents all the answer tokens preceding the current prediction token $x_{i}$, where $i$ denotes the step in the sequence of text token generation. $\mathcal{F}_{\mathrm{t}} \in \mathbb{R}^{N_{\mathrm{t}} \times d_{\mathrm{t}}}$ is the token embedding of the input question, $N_{\mathrm{t}}$ and $d_{\mathrm{t}}$ denote the number of text tokens and token embedding dimension. By infusing these enriched visual cues into the training pipeline, MLLMs can develop a more comprehensive understanding of visual content, leading to better alignment between visual and textual representations. To accelerate the training process, we follow LoRA Augmented Training (LAF) strategy ~\cite{jiao2024enhancing} to perform fine-tuning on Mipha-3B~\cite{zhu2024comprehensive} and LLaVA-1.5~\cite{liu2023improved} using LoRA ~\cite{hu2021lora}.
\section{Experiment}
In this section, we conduct a comprehensive comparison of our method with existing state-of-the-art (SOTA) multimodal models. Additionally, we perform a series of ablation studies to further validate the proposed method. Finally, we provide visualization examples for in-depth analysis. 

\noindent\textbf{Models.} For the vision encoder, we adopt SigLIP-384px~\cite{zhai2023sigmoid} for experiments. We leverage Phi-2-2.7B~\cite{microsoft2024phi2} and Vicuna-7B~\cite{vicuna} model as the language decoder. For the multimodal projector, same as LLaVA~\cite{liu2023improved}, we adopt a two-layer MLP. We use OpenSeed~\cite{zhang2023simple} and PaddleOCRv2~\cite{du2021pp} to generate the per-pixel externally knowledge for pixel class and OCR text, and leverage UAE-Large-V1~\cite{li2023angle} to extract the textual embedding. 

\noindent\textbf{Training Setting.} We fine-tune the models on LLaVA-Instruct-150K dataset~\cite{liu2023improved} using LoRA~\cite{hu2021lora} for 1 epoch, at a learning rate of 2e-4 and a batch size of 256 on 32 $\times$ V100 32GB GPUs. For the setting of LoRA, we set LoRA rank to be 128 and LoRA's hyperparameter $\alpha$ as 256. Note that we fix all the weights of pre-trained modules, \textit{i.e.}, vision encoder, language encoder and MLP, during training. Our models' weights are initialized from Mipha-3B~\cite{zhu2024comprehensive} and LLava-7B~\cite{liu2023improved}.

\noindent\textbf{Benchmarks and Baselines.}We evaluate our approach using $9$ popular benchmarks to comprehensively assess its multimodal capabilities. These benchmarks include: VQA-v2 test-dev split~\cite{goyal2017making}, GQA test-dev-balanced split~\cite{hudson2019gqa}, ScienceQA-IMG test split~\cite{lu2022learn}, MME perception~\cite{mme}, MME cognition~\cite{mme}, MMBench test split~\cite{liu2023mmbench}, MM-Vet test split~\cite{yu2023mm}, TextVQA~\cite{singh2019towards}, and POPE~\cite{li2023evaluating}. 

We compare our results with a bunch of state-of-the-art multimodal large language models (MLLMs): BLIP-2~\cite{li2023blip}, InstructBLIP~\cite{dai2024instructblip}, Shikra-13B~\cite{chen2023shikra}, IDEFICS80/9B~\cite{laurenccon2024obelics}, Qwen-VL~\cite{bai2023qwen}, mPLUG-Owl2~\cite{ye2023mplug}, LLaVA-v1.5-13/7B~\cite{liu2023improved}, LAF-7B~\cite{jiao2024enhancing}, and multimodal small language models (MSLMs) ~\cite{zhu2024comprehensive}: MobileVLM~\cite{chu2024mobilevlm}, LLaVA-Phi~\cite{zhu2024llava}, MC-LLaVA~\cite{MC-LLaVA}, Imp-v1~\cite{sung2023empirical}, MoE-LLaVA-3.6B~\cite{lin2024moe}, TinyLLaVA-share-Sig-Phi~\cite{zhou2024tinyllava}, Bunny~\cite{he2024efficient} and Mipha~\cite{zhu2024comprehensive}.

\begin{table*}[!t]
  \centering
  \footnotesize
  \caption{The ablation study of different visual prompts. \textit{Mipha-3B} is the baseline with standard visual \& text prompt. \textit{Mipha-3B+LAF} denotes using textual prompting with LoRA Augmented Fine-tuning following (LAF)~\cite{jiao2024enhancing}. \textit{feature fusion} and \textit{feature addition} represent two prompt fusion methods we use to insert the auxiliary visual prompt to the original image features.} 
  \label{tbl:full_table}
  \resizebox{1.0\linewidth}{!}{
      \begin{tabular}{l|cccccccccc}
        \toprule
           Method  & $\text{VQAv2}$ &\text{GQA} & \text{SQA$^{\text{I}}$} & $\text{VQA}^{\text{T}}$ & \text{MME-P} & \text{MME-C} & \text{MMB} & \text{MM-Vet} & \text{POPE} \\
        \midrule
        Mipha-3B & 81.3 & 63.9 & 70.9 & 56.6 & 1488.9 & 295.0 & 69.7 & 32.1 & 86.7\\
        Mipha-3B + LAF  & 81.6{$\greenup$}   &62.6{$\redown$}   & 71.4{$\greenup$}   & 57.8{$\greenup$}   & 1472.3{$\redown$}  & 356.8{$\greenup$}  & 71.0{$\greenup$}  & 34.8{$\greenup$}   & 88.5{$\greenup$}  \\
        Ours (feature fusion)  &  81.9{$\greenup$} & 64.8{$\greenup$} & 71.6{$\greenup$} & 57.6{$\greenup$} & 1493.5{$\greenup$} & 345.5{$\greenup$} & 71.3{$\greenup$} & 34.3{$\greenup$} &88.5{$\greenup$}  \\
        \rowcolor{LightCyan}Ours (feature addition)  &  \textbf{82.4}{$\greenup$} & \textbf{65.3}{$\greenup$} & \textbf{71.8}{$\greenup$} & \textbf{57.8}{$\greenup$} & \textbf{1501.2}{$\greenup$} & \textbf{369.1}{$\greenup$} & \textbf{71.5}{$\greenup$} & \textbf{35.1}{$\greenup$} & \textbf{88.7}{$\greenup$}\\
        \bottomrule
      \end{tabular}
    }
\label{table2}
\vspace{-2mm}
\end{table*}

\begin{table*}[!t]
  \centering
  \footnotesize
  \caption{The ablation study of using different vision encoders, i.e., SigLIP v.s. CLIP. The results of \textit{Mipha-3B} on CLIP are from ~\cite{zhu2024comprehensive}.}
  \label{tbl:full_table}
  \resizebox{1.0\linewidth}{!}{
      \begin{tabular}{l|c|ccccccccc}
        \toprule
           Method & Vis Enc & $\text{VQAv2}$ &\text{GQA} & \text{SQA$^{\text{I}}$} & $\text{VQA}^{\text{T}}$ & \text{MME-P} & \text{MME-C} & \text{MMB} & \text{MM-Vet} & \text{POPE} \\
        \midrule
        Mipha-3B & CLIP& 78.6 & 62.3 &  68.2 & 53.0 & - & - & 68.4 & 31.0& 86.9\\
        \rowcolor{LightCyan}Ours & CLIP & \textbf{79.7}{$\greenup$}  & \textbf{63.7}{$\greenup$} & \textbf{70.1}{$\greenup$} & \textbf{54.8}{$\greenup$} & \textbf{1445.5} & \textbf{308.4} & \textbf{70.1}{$\greenup$} & \textbf{33.7}{$\greenup$} & \textbf{88.8}{$\greenup$} \\
        \midrule
        Mipha-3B & SigLIP & 81.3& 63.9 & 70.9 & 56.6 & 1488.9 & 295.0 & 69.7 & 32.1 & 86.7\\
        \rowcolor{LightCyan}Ours & SigLIP &  \textbf{82.4}{$\greenup$} & \textbf{65.3}{$\greenup$} & \textbf{71.8}{$\greenup$} & \textbf{57.8}{$\greenup$} & \textbf{1501.2}{$\greenup$} & \textbf{369.1}{$\greenup$} & \textbf{71.5}{$\greenup$} & \textbf{35.1}{$\greenup$} & \textbf{88.7}{$\greenup$}\\
        \bottomrule
      \end{tabular}
    }
\label{table3}
\vspace{-1mm}
\end{table*}

\begin{table}
  \begin{minipage}{0.99\textwidth}
\centering  
\vspace{-4mm}
\scalebox{0.86}{
\begin{tabular}{c p{6.6cm} p{6.6cm} }
\toprule
 \multicolumn{3}{l}{\bf Visual input example, context-related questions:}  \\
\midrule
&  \includegraphics[height=4.5cm,width=6.6cm]{figures/figure4_1.pdf} &  \includegraphics[height=4.5cm,width=6.6cm]{figures/figure4_2.pdf} \\
& What is funny about this image? & {Why are the persons wearing black clothes?}
\\ \midrule
LLaVA-7B & The funny aspect of this image is the juxtaposition of two women pointing fingers at a cat, which is sitting at a table with a plate of food. The cat appears to be looking at the women with an angry expression, as if it is reacting to their pointing fingers. This creates a humorous and unexpected scene, as it is not common to see a cat reacting to human gestures in such a manner. The image captures a light-hearted and amusing moment between the women and the cat. & The persons in the image are wearing black clothes, which is a common attire for religious ceremonies or events. The group of people, including a man and a woman, are walking down the street, possibly participating in a religious procession or a funeral. The black clothing signifies their respect and adherence to the religious or cultural traditions associated with the event they are attending.
\\ \midrule
\textbf{Ours} & The funny aspect of this image is the contrast between the two women pointing at the cat and the cat's reaction to the situation. \textcolor{brown}{One woman is making a funny face, while the cat is sitting on a table with a plate of food, seemingly unbothered by the attention. The cat's calm demeanor in the face of the women's playful behavior creates a humorous scene, as the women try to entertain themselves and the cat seems to be enjoying the attention without reacting to their attempts.} & The persons in the image are wearing black clothes, \textcolor{brown}{which are typically associated with mourning or a period of grieving. This suggests that they might be attending a funeral or a memorial service for someone who has passed away.} The black attire signifies respect and solidarity with the bereaved family and friends during this difficult time. \textcolor{brown}{It is a common practice in many cultures to wear black clothes to express sympathy and support for those persons who have experienced a loss.}
\\ \bottomrule
\end{tabular}
}
\vspace{1mm}
\captionof{table}{The challenging examples on LLaVA-1.5, our approach can provide reasonable reponses for the tricky questions which need to fully understand the contexts of the images.}
\label{tab:example}  
  \end{minipage}
\vspace{-4mm}
\end{table}

\begin{table*}[!t]
  \centering
  \caption{The ablation study of introducing OCR information into the visual prompt.} 
  \vspace{-2mm}
  \label{tbl:full_table}
  \resizebox{1.0\linewidth}{!}{
      \begin{tabular}{l|ccccccccc}
        \toprule
           Method  & $\text{VQAv2}$ &\text{GQA} & \text{SQA$^{\text{I}}$} & $\text{VQA}^{\text{T}}$ & \text{MME-P} & \text{MME-C} & \text{MMB} & \text{MM-Vet} & \text{POPE} \\
        \midrule
        Mipha-3B & 81.3 & 63.9 & 70.9 & 56.6 & 1488.9 & 295.0 & 69.7 & 32.1 & 86.7\\
        Ours (w/o OCR) &81.9{$\greenup$} & 64.7{$\greenup$}  & 71.3{$\greenup$} & 57.1{$\greenup$} & 1498.3{$\greenup$} & 355.2{$\greenup$} & 70.8{$\greenup$}  &  34.0{$\greenup$} & 87.9{$\greenup$} \\
        \rowcolor{LightCyan}Ours  &  \textbf{82.4}{$\greenup$} & \textbf{65.3}{$\greenup$} & \textbf{71.8}{$\greenup$} & \textbf{57.8}{$\greenup$} & \textbf{1501.2}{$\greenup$} & \textbf{369.1}{$\greenup$} & \textbf{71.5}{$\greenup$} & \textbf{35.1}{$\greenup$} & \textbf{88.7}{$\greenup$}\\
        \bottomrule
      \end{tabular}
    }
\label{table5}
\vspace{-2mm}
\end{table*}

\subsection{Ablation Studies}
\label{ablation}
In this section, we conduct an ablation study to assess the effectiveness of the proposed approach. By default, the experiments are conducted using Mipha-3B~\cite{zhu2024comprehensive} with Phi-2~\cite{microsoft2024phi2} as the language backbone unless otherwise specified.

\paragraph{Prompting MLLMs with Different Approaches.}
In Table \ref{table2}, we present the results of the ablation study for four different prompting strategies:
(1) Mihpa-3B baselines with vanilla text prompt, as used by LLaVA-1.5~\cite{liu2023improved}.
(2) Mihpa-3B + LAF proposed in~\cite{jiao2024enhancing} that appends externel local contextual knowledge to the text prompts.
(3) The proposed auxiliary visual prompt inserted via feature fusion.
(4) The proposed auxiliary visual prompt  added via feature addition.

From Table \ref{table2}, we note that compared to the baseline (1) with vanilla prompts, both proposed fusion strategies (3) and (4) exhibit a significant improvement. This suggests that external knowledge is indeed beneficial in enhancing the capabilities of MLLMs. In comparison to Mihpa-3B+LAF (2), which inserts external local contextual knowledge into the text prompt, (4) outperforms it in 8 out of 9 benchmarks, notably for GQA~\cite{hudson2019gqa} and MME-P~\cite{mme}. This implies that explicitly linking external local knowledge to the original visual features reduces the model's learning burden in establishing spatial relationships, consequently enhancing performance. Furthermore, we empirically observe that directly adding auxiliary visual prompts yields slightly better results than concatenation. Therefore, we adopt feature addition as our default setting for subsequent experiments.

\paragraph{The Effect of Using Different Vision Encoders. } In Table \ref{table3}, we further ablate the effectiveness brought by different vision encoders, i.e., CLIP~\cite{radford2021learning} v.s. SigLIP~\cite{zhai2023sigmoid}. From the results, we can draw two conclusions. First, for both vision encoders, our methods have consistent improvement compared to the baselines, which validates the stability of our methods. Second, SigLIP emerges as the stronger vision encoder when compared to CLIP. Therefore, we opt to utilize SigLIP as the default vision encoder in subsequent sections.

\paragraph{The Effect of Introducing OCR Information Into Visual Prompt. } In Table \ref{table5}, we perform the ablation of using OCR information or not, we can conclude that adopting the information from OCR can further improve the model's overall performance, especially, when incorporating OCR information for some text-specific tasks (\textit{e.g.}, TextVQA~\cite{singh2019towards} and MM-Vet~\cite{yu2023mm}), it can achieve remarkable performance boosts. 

\begin{table*}[!t]
  \centering
  \caption{The comprehensive multi-modal evaluation across $9$ distinct benchmarks to thoroughly assess model performance: $\text{VQA}{\text{v2}}$~\cite{goyal2017making}, GQA~\cite{hudson2019gqa}, $\text{SQA}^{\text{I}}$: ScienceQA-IMG~\cite{lu2022learn}, $\text{VQA}^{\text{T}}$: TextVQA~\cite{singh2019towards}, \text{MME-P}: MME Perception~\cite{mme}, \text{MME-C}: MME Cognition~\cite{mme}, \text{MMB}: MMBench~\cite{liu2023mmbench}, \text{MM-Vet}~\cite{yu2023mm}, and \text{POPE}~\cite{li2023evaluating}. The included proprietary in-house data not publicly accessible, denoted as {$^\dagger$}. The image resolution used by the visual backbone is indicated in the column labeled \textit{Res.}, while the columns \textit{PT} and \textit{IT} represent the data sizes in the pretraining and visual instruction tuning stages, respectively.} 
  \label{tbl:full_table}
  \resizebox{1.0\linewidth}{!}{
      \begin{tabular}{l|l|l|l|l|ccccccccc}
        \toprule
           Method & LM  & Res. & PT & IT & $\text{VQAv2}$ &\text{GQA} & \text{SQA$^{\text{I}}$} & $\text{VQA}^{\text{T}}$ & \text{MME-P} & \text{MME-C} & \text{MMB} & \text{MM-Vet} & \text{POPE} \\
        \midrule
        \multicolumn{14}{c}{Multimodal Large Language Models} \\
        \midrule
        BLIP-2~\cite{li2023blip} & Vicuna (13B) & 224 & 129M & -                               & 65.0 & 41.0 & 61.0 & 42.5 & 1293.8 & 290.0 & - & 22.4 & 85.3\\
        InstructBLIP~\cite{dai2024instructblip} & Vicuna (7B) & 224 & 129M & 1.2M                 & - & 49.2 & 60.5 & 50.1 & - & - & 36 & 26.2 & - \\
        InstructBLIP~\cite{dai2024instructblip} & Vicuna (13B) & 224 & 129M & 1.2M                & - & 49.5 & 63.1 & 50.7 & 1212.8 & 291.8 & - & 25.6 & 78.9 \\
        Shikra~\cite{chen2023shikra} & Vicuna (13B) & 224 & 600K & 5.5M                            &  77.4 & - & - & - & - & - & 58.8 & - & -\\
        IDEFICS-9B~\cite{laurenccon2024obelics}& LLaMA (7B) & 224 & 353M & 1M                           & 50.9 & 38.4 & - & 25.9 & - & - & 48.2 & - & -  \\
        IDEFICS-80B~\cite{laurenccon2024obelics}& LLaMA (65B) & 224 & 353M & 1M                         & 60.0 & 45.2 & - & 30.9 & - & - & 54.5 & - & - \\
        Qwen-VL~\cite{bai2023qwen}& Qwen (7B) & 448 & 1.4B{$^\dagger$} & 50M{$^\dagger$}         & 78.8 & 59.3 & 67.1 & \textbf{63.8} & - & - & 38.2 & - & -\\
        Qwen-VL-Chat~\cite{bai2023qwen} & Qwen (7B) & 448 & 1.4B{$^\dagger$} & 50M{$^\dagger$}    & 78.2 & 57.5 & 68.2 & 61.5 & 1487.5 & 360.7 & 60.6 & - & -\\
        mPLUG-Owl2~\cite{ye2023mplug} & LLaMA (7B) & 448 & 400M & 1.23M                      & 79.4 & 56.1 & 68.7 & 58.2 & 1450.2 & 313.2 & 64.5 & \textbf{36.2} & 85.8 \\
        LLaVA-1.5~\cite{liu2023improved} & Vicuna (7B) & 336 & 558K & 665K                        & 78.5 & 62.0 & 66.8 & 58.2 & 1510.7 & 316.1 & 64.3 & 30.5 & 85.9 \\
         LAF-7B~\cite{jiao2024enhancing} & Vicuna (7B) & 336 & 558K & 665K                       & 79.0 & 60.5 & - & 60.1 & 1482.7 & 397.9 & 67.3 & 35.2 & 88.9\\ 
        \midrule
        \rowcolor{LightCyan} \textbf{LLaVA-1.5{$^+$}(Ours)} & Vicuna(7B)  & 336 & 558K & 665K  & 79.8$\greenup$ & 63.3$\greenup$  & 69.5$\greenup$ & 59.8$\greenup$ & \textbf{1515.3}$\greenup$ & \textbf{399.5}$\greenup$ & 67.6$\greenup$  & 34.9$\greenup$  & \textbf{88.9}$\greenup$ \\
        \midrule
        \multicolumn{14}{c}{Multimodal Small Language Models} \\
        \midrule

        MobileVLM-1.7B~\cite{chu2023mobilevlm} & M-LLaMA (1.4B) & 336 & 558K & 665K & - & 56.1 &  57.3 & 41.5 & 1196.2 & - & 53.2 & - & 84.5 \\ 
        MobileVLM-3B ~\cite{chu2023mobilevlm}& M-LLaMA (2.7B) & 336 & 558K & 665K & - & 59.0 & 61.2 & 47.5 & 1288.9 & - & 59.6 & - & 84.9 \\
        MobileVLM-v2-1.7B~\cite{chu2024mobilevlm} & M-LLaMA (1.4B) & 336 & 1.2M & 2.4M & -  & 59.3 & 66.7 & 52.1 & 1302.8 & - & 57.7 & - & 84.3\\
        MobileVLM-v2-3B~\cite{chu2024mobilevlm} & M-LLaMA (2.7B) & 336 & 1.2M & 2.4M & -  & 61.1 & 70.0 & 57.5 & 1440.5 & - & 63.2 & - & 84.7 \\
        LLaVA-Phi~\cite{zhu2024llava} & Phi-2 (2.7B) & 336 & 558k & 665K  & 71.4 & -  & 68.4 & 48.6 & 1335.1 & - & 59.8 & 28.9 & 85.0 \\
        MC-LLaVA~\cite{MC-LLaVA} & Phi-2 (2.7B) & 384 & 558k & 665K  & 64.2 & 49.6  & - &  38.6 & - & - & - & - & 80.6\\
        Imp-v1~\cite{sung2023empirical} & Phi-2 (2.7B) & 384 & 558K & 665K  & 79.5 & 58.6  & 70.0 & 59.4 & 1434.0 & - & 66.5 & 33.1 & 88.0 \\
        MoE-LLaVA-3.6B~\cite{lin2024moe} & Phi-2 (2.7B) & 384 & 558k & 1.59M & 79.9 & 62.6 & 70.3 & 57.0 & 1431.3 & - & 68.0 & 35.9 & 85.7 \\
        TinyLLaVA~\cite{zhou2024tinyllava} & Phi-2 (2.7B) & 384 & 1.2M & 665k & 79.9 & 62.0 & 69.1 & 59.1 & 1464.9 & - & 66.9 & 32.0 & 86.4 \\
        Bunny-3B~\cite{he2024efficient} & Phi-2 (2.7B) & 384 & 2M & 695K & 79.8 & 62.5 & 70.9 & - & 1488.8 & 289.3 & 68.6 & - & 86.8\\
        Mipha-3B~\cite{zhu2024comprehensive} & Phi-2 (2.7B)  & 384 & 558K & 665K & 81.3 & 63.9 & 70.9 & 56.6 & 1488.9 & 295.0 & 69.7 & 32.1 & 86.7\\
        \midrule
        \rowcolor{LightCyan} \textbf{Mipha-3B{$^+$}(Ours)} & Phi-2 (2.7B)  & 384 & 558K & 665K  &  \textbf{82.4}$\greenup$ & \textbf{65.3}$\greenup$ & \textbf{71.8}$\greenup$ & 57.8$\greenup$ & 1501.2$\greenup$ & 369.1$\greenup$ & \textbf{71.5}$\greenup$ & 35.1$\greenup$ & 88.7$\greenup$\\
        \bottomrule
      \end{tabular}
    }
\label{table1}
\end{table*}

\subsection{Main Results}

In Table~\ref{table1}, we compare our methods with other state-of-the-art (SOTA) models. We divide the table into sections for language models smaller than $3$B and those beyond $7$B to provide a clearer comparison. From the results, we observe that our model achieves the best performance on $7$ out of $9$ benchmarks for larger language models (>7B) and attains the highest accuracy on $7$ out of $9$ benchmarks for relatively smaller language models (<3B). Note that, in Table~\ref{table1}, some models, e.g. Shikra-13B~\cite{chen2023shikra}, Qwen-VL~\cite{bai2023qwen}, are trained with million or billion level data, while our model is only trained on the dataset used by LLaVA-1.5 without any extra data for neither pre-training nor fine-tuning, which highlights the exceptional multimodal understanding and reasoning capabilities of our models. In addition, on top of the LLaVA-1.5 framework, our approach can bring more remarkable and consistent improvement on all benchmarks compared with LAF~\cite{jiao2024enhancing}. It justifies the proposed infusion strategy, which involves inserting external knowledge in a pixel-wise manner directly into the visual features, as being more effective than appending it to the text prompt~\cite{jiao2024enhancing}.

\subsection{Quantitative Result Analysis}
We present visualization results in Table \ref{tab:example} and \ref{tab:example2} to further illustrate the improvement of our model in terms of both global image understanding and local object and text recognition.
Table \ref{tab:example} demonstrates that compared to LLaVA-1.5 7B~\cite{liu2023improved}, our approach generates more detailed and contextually relevant responses, \textit{e.g.}, ``The cat’s calm demeanor in the face
of the women’s playful behavior'' for the left example; ``mourning or a period of grieving'' and ``express sympathy and support for those persons who have experienced a loss'' for the right example, which all need a deeper understanding of the global image context.
Meanwhile, Table \ref{tab:example2} highlights our model's ability to correctly recognize objects' spatial relationships, such as between a ``desk lamp'' and a ``laptop'' from the left image, and exhibit stronger OCR capability in detecting words written on a book from the right image, compared to LLaVA-1.5 7B~\cite{liu2023improved}.
These visualizations validate the effectiveness of our proposed methods and support the conclusion that incorporating external local contextual information in a spatial-wise manner improves the model's fine-grained recognition capability and enhances its overall ability for global image understanding. Note that we've shown more ablation study experiments and visualization result analysis in the Appendix.

\begin{table}
  \begin{minipage}{0.99\textwidth}
\centering  
\vspace{-4mm}
\scalebox{0.86}{
\begin{tabular}{c p{6.4cm} p{6.4cm} }
\toprule
 \multicolumn{3}{l}{\bf Visual input example, relationship-aware and text-related questions:}  \\
\midrule
&  \includegraphics[height=4.5cm,width=6.4cm]{figures/figure5_1.pdf} &  \includegraphics[height=4.5cm,width=6.4cm]{figures/figure5_2.pdf} \\
& On the right desk, what is to the left of the laptop? & What are all the scene text in the image?
\\ \midrule
LLaVA-1.5 7B & To the left of the laptop, there is a stack of books on the desk.  &  The scene text in the image is "Shakespeare's Dramas, Sonnets, \& Poems".
\\ \midrule
\textbf{Ours} & There is a \textcolor{brown}{desk lamp} to the left of the laptop on the right desk. & The scene text in the image includes the title \textcolor{brown}{"Shakespeare's Comedies, Histories, and Tragedies"}.
\\ \bottomrule
\end{tabular}
}
\vspace{1mm}
\captionof{table}{The challenging examples on LLaVA-1.5. Our approach can generate accurate responses for text-related questions.}
\vspace{-5mm}
\label{tab:example2}  
  \end{minipage}
\end{table}

\section{Limitations and Broader Impact}
Our method relies on pre-trained models for panoptic segmentation and OCR detection in a zero-shot manner. The performance of these models will significantly impact the performance of our proposed method, particularly when there is a substantial domain gap between the images from specific benchmarks and the training set of the segmentation or OCR models. 
While the proposed approach holds promise for significantly enhancing the cognitive capabilities of multimodal models and may inspire new methodologies and techniques in the development of robust multimodal AI systems, users must be aware of potential negative societal impacts. For instance, biases may manifest in various forms; for example, biased responses may be generated by the model if the training data of MLLMs, panoptic segmentation, and OCR detection models contain certain biases.
\section{Conclusion}
In this paper, we have proposed a method for leveraging external knowledge, such as localized contextual information, to enhance the capabilities of multimodal language models (MLLMs). To accomplish this objective, we propose extracting pixel-wise contextual information using a panoptic segmentation and OCR model, and then directly integrate this with the visual features. This enables the model to better understand both fine-grained objects and the overall global image context. Experimental results from ablations and comparisons with state-of-the-art methods demonstrate the effectiveness of our approach. We hope this paper can shed light on the importance of external knowledge for MLLMs and an effective way to leverage such knowledge.

\bibliographystyle{plainnat}
\bibliography{neurips_2024}

\newpage
\appendix  
\section{Appendix}
In the supplementary materials, we provide the following sections:

\begin{enumerate}[label=(\alph*)]
\item More implementation details in Section \ref{implementation}.
 
\item Ablation study experiments in Section \ref{ablation}.
   
\item Visualization result analysis in Section \ref{visualization}.

\end{enumerate}

\section{Implementation Details}
\label{implementation}
The training time for LLaVA-1.5 7B~\cite{liu2023improved} and Mipha-3B~\cite{zhu2024comprehensive} is approximately 14 hours and 9 hours, respectively, with a batch size of 256 on 32 × NVIDIA V100 32GB GPUs. For the initialization of the proposed prompt embedding network (PEN), we use Kaiming initialization technology~\cite{he2015delving}. The UAE-Large-V1\footnote{\url{https://huggingface.co/WhereIsAI/UAE-Large-V1}} model is adopted as the pre-trained textual encoder to extract textual embeddings for the visual prompt.

\section{Ablation Study}
\label{ablation}

Next, we conduct more ablation study experiments to provide deeper insight into the components of our proposed approach.

\paragraph{The Effect of Using Different Pre-trained Textual Encoders. } In Table \ref{sup_table1}, we perform an ablation study using different textual encoders, \textit{i.e.}, CLIP~\cite{radford2021learning} vs. UAE~\cite{li2023angle}, to extract textual embeddings for the proposed visual prompt. We draw two conclusions from Table~\ref{sup_table1}: (1) Using different textual encoders, the proposed approach consistently outperforms the baseline, demonstrating the robustness of our method. (2) Adopting UAE as the pre-trained textual encoder achieves significantly better performance. Therefore, we choose UAE as the default pre-trained textual encoder in our experiments.

\paragraph{Object Detector v.s. Segmentation Model. } To determine the effect of using an object detector or segmentation model to incorporate pixel-level semantics into the proposed visual prompt, we conduct an ablation study with the popular object detector GroundingDINO~\cite{liu2023grounding} and the segmentation model OpenSeed~\cite{zhang2023simple}. The results are shown in Table \ref{sup_table2}. We observe that both GroundingDINO and OpenSeed significantly boost performance across all benchmarks. However, utilizing OpenSeed achieves better performance gains due to its fine-grained mask regions. Thus, we adopt OpenSeed by default to generate object regions.

\paragraph{The Effect of Fine-Tuning with the Visual Prompt. } As displayed in Table \ref{sup_table3}, the model fine-tuned with the proposed visual prompt (\textit{i.e.}, the third row) achieves remarkably better performance than the one fine-tuned without our visual prompt (\textit{i.e.}, the second row) across all benchmarks. Specifically, without using our visual prompt for fine-tuning, the model even shows performance degradation on Text-VQA benchmark~\cite{singh2019towards} and has negligible gains on Science-QA~\cite{lu2022learn}, VQAv2~\cite{goyal2017making}, MME-P~\cite{mme}, and MME-C~\cite{mme} benchmarks. All these results demonstrate the superiority of the proposed method.

\section{Visualization Result Analysis}
\label{visualization}
We've provided more visualization results in Table \ref{sup_tab:example1}, \ref{sup_tab:example2}, \ref{sup_tab:example3}, and \ref{sup_tab:example4}. Compared to LLaVA-1.5 7B~\cite{liu2023improved}, our method generates more reasonable and accurate responses to the questions.

As shown in Table \ref{sup_tab:example1}, our approach can generate accurate movie titles, such as ``The Godfather'', and the two actors' names, such as ``Al Pacino'' and 'Robert De Niro''. Additionally, it provides a corresponding introduction, such as ``The movie is a classic crime drama film directed by Francis Ford Coppola, known for its iconic characters, storytelling, and memorable scenes" for the left example. In the right example, our method generates the precise title ``The Lord of the Rings: The Fellowship of the Ring'' and provides an accurate introduction, such as ``The movie is an epic fantasy adventure that follows the journey of a fellowship of characters". In contrast, LLaVA's responses are very general without fully understanding the global contexts within the images.

In Table \ref{sup_tab:example2}, our method not only recognizes the text ``Me: I’ll do it at 8. Time: 8.05. Looks like I gotta wait till 9 now'', but also understands its implication, such as ``the puppet is aware of the time and is intentionally delaying any work or task it might be assigned'', in the left image. For the right image, our method generates more potential options, such as a ``private school'', ``a religious school'', etc. In contrast, LLaVA's answers are either inappropriate or too limited.

Finally, as shown in Table \ref{sup_tab:example3} and \ref{sup_tab:example4}, our method understands spatial relationships and accurately recognizes the text within the images. For example, in the left example of Table \ref{sup_tab:example3}, our approach accurately names the person on the left as ``Keira Knightley''. It also recognizes the plate number ``S3302CD'' in the left example of Table \ref{sup_tab:example4}, while LLaVA's answers are all incorrect. This indicates the strong fine-grained multimodal understanding capacity of our proposed method.

\begin{table*}[t]
  \centering
  \footnotesize
  \caption{The ablation study of using different textual encoders, i.e., CLIP v.s. UAE, to extract textual embeddings for the proposed visual prompt.}
  \label{tbl:full_table}
  \resizebox{1.0\linewidth}{!}{
      \begin{tabular}{l|c|ccccccccc}
        \toprule
           Method & Text Enc & $\text{VQAv2}$ &\text{GQA} & \text{SQA$^{\text{I}}$} & $\text{VQA}^{\text{T}}$ & \text{MME-P} & \text{MME-C} & \text{MMB} & \text{MM-Vet} & \text{POPE} \\
        \midrule
        Mipha-3B & - & 81.3& 63.9 & 70.9 & 56.6 & 1488.9 & 295.0 & 69.7 & 32.1 & 86.7\\
        Mipha-3B & CLIP & 82.1{$\greenup$} & 64.9{$\greenup$} & 71.3{$\greenup$} & 57.4{$\greenup$} & 1497.2{$\greenup$} & 361.5{$\greenup$} & 71.1{$\greenup$} & 34.6{$\greenup$} & 88.5{$\greenup$}\\
        \rowcolor{LightCyan}Ours & UAE &  \textbf{82.4}{$\greenup$} & \textbf{65.3}{$\greenup$} & \textbf{71.8}{$\greenup$} & \textbf{57.8}{$\greenup$} & \textbf{1501.2}{$\greenup$} & \textbf{369.1}{$\greenup$} & \textbf{71.5}{$\greenup$} & \textbf{35.1}{$\greenup$} & \textbf{88.7}{$\greenup$}\\
        \bottomrule
      \end{tabular}
    }
\label{sup_table1}
\end{table*}

\begin{table*}[t]
  \centering
  \footnotesize
  \caption{The ablation study of using an object detector or a panoptic segmentation model to extract object regions for pixel-level textual embeddings.}
  \label{tbl:full_table}
  \resizebox{1.0\linewidth}{!}{
      \begin{tabular}{l|c|ccccccccc}
        \toprule
           Method & Region Generator & $\text{VQAv2}$ &\text{GQA} & \text{SQA$^{\text{I}}$} & $\text{VQA}^{\text{T}}$ & \text{MME-P} & \text{MME-C} & \text{MMB} & \text{MM-Vet} & \text{POPE} \\
        \midrule
        Mipha-3B & - & 81.3& 63.9 & 70.9 & 56.6 & 1488.9 & 295.0 & 69.7 & 32.1 & 86.7\\
        Mipha-3B & GroundingDINO & 82.0{$\greenup$}  & 64.9{$\greenup$}  & 71.4{$\greenup$}  & 57.2{$\greenup$}  & 1491.7{$\greenup$}  &  350.2{$\greenup$}  & 71.0{$\greenup$}   & 34.5{$\greenup$}  & 88.4{$\greenup$}  \\
        \rowcolor{LightCyan}Ours & OpenSeed &  \textbf{82.4}{$\greenup$} & \textbf{65.3}{$\greenup$} & \textbf{71.8}{$\greenup$} & \textbf{57.8}{$\greenup$} & \textbf{1501.2}{$\greenup$} & \textbf{369.1}{$\greenup$} & \textbf{71.5}{$\greenup$} & \textbf{35.1}{$\greenup$} & \textbf{88.7}{$\greenup$}\\
        \bottomrule
      \end{tabular}
    }
\label{sup_table2}
\end{table*}

\begin{table*}[!t]
  \centering
  \footnotesize
  \caption{The ablation study of fine-tuning with and without the proposed visual prompt. The first, second and third rows mean Mipha-3B baseline, fine-tuning on Mipha-3B without and with the proposed visual prompt using LoRA~\cite{hu2021lora}.}
  \label{tbl:full_table}
  \resizebox{1.0\linewidth}{!}{
      \begin{tabular}{l|c|ccccccccc}
        \toprule
           Method & Visual Prompt & $\text{VQAv2}$ &\text{GQA} & \text{SQA$^{\text{I}}$} & $\text{VQA}^{\text{T}}$ & \text{MME-P} & \text{MME-C} & \text{MMB} & \text{MM-Vet} & \text{POPE} \\
        \midrule
        Mipha-3B & - & 81.3& 63.9 & 70.9 & 56.6 & 1488.9 & 295.0 & 69.7 & 32.1 & 86.7\\
        Mipha-3B$^{+}$ & \xmark & 81.4{$\greenup$}  & 64.3{$\greenup$}  & 71.0{$\greenup$}  & 56.5{$\redown$}  & 1489.2{$\greenup$}  &  303.2{$\greenup$}  & 70.4{$\greenup$}   & 33.5{$\greenup$}  & 87.4{$\greenup$}  \\
        \rowcolor{LightCyan}Ours & \cmark &  \textbf{82.4}{$\greenup$} & \textbf{65.3}{$\greenup$} & \textbf{71.8}{$\greenup$} & \textbf{57.8}{$\greenup$} & \textbf{1501.2}{$\greenup$} & \textbf{369.1}{$\greenup$} & \textbf{71.5}{$\greenup$} & \textbf{35.1}{$\greenup$} & \textbf{88.7}{$\greenup$}\\
        \bottomrule
      \end{tabular}
    }
\label{sup_table3}
\end{table*}

\clearpage
\begin{table}
  \begin{minipage}{0.99\textwidth}
\centering  
\vspace{-4mm}
\scalebox{0.9}{
\begin{tabular}{c p{6.2cm} p{6.2cm} }
\toprule
 \multicolumn{3}{l}{\bf Visual input example, context-related questions:}  \\
\midrule
&  \includegraphics[height=4.5cm,width=6.2cm]{figures/sup_figure1_1.pdf} &  \includegraphics[height=4.5cm,width=6.2cm]{figures/sup_figure1_2.pdf} \\
& Can you give a short introduction to this movie? & {Can you give a short introduction to this movie?}
\\ \midrule
LLaVA-7B & The image features a scene from a movie where two men are standing close to each other, with one man placing his hand on the other's shoulder. They are both wearing tuxedos, which suggests that they are attending a formal event or a special occasion. The men appear to be engaged in a conversation or interacting with each other in some way, possibly discussing the event or sharing a moment of camaraderie. & The image features a young man wearing a brown shirt and holding a ring in his hand. He appears to be in a dark setting, possibly a cave or a dungeon. The man is looking at the camera, possibly engaging in a conversation or sharing a story. The ring he is holding suggests that the scene might be related to a fantasy or adventure movie, where the ring could represent a magical or powerful object.
\\ \midrule
\textbf{Ours} & The image features a scene from the movie \textcolor{brown}{``The Godfather'', where two men played by actors Al Pacino and Robert De Niro, are standing close to each other.} One man has his hand on the other's shoulder, and they appear to be engaged in a conversation or a confrontation. \textcolor{brown}{The movie is a classic crime drama film directed by Francis Ford Coppola, known for its iconic characters, storytelling, and memorable scenes.} & The image features a scene from the movie \textcolor{brown}{``The Lord of the Rings: The Fellowship of the Ring'', where the character Gandalf is holding out his hand to the hobbit Frodo, offering him the One Ring.} This iconic moment captures the essence of the story, where Frodo is entrusted with the task of destroying the ring to prevent its power from falling into the wrong hands. \textcolor{brown}{The movie is an epic fantasy adventure that follows the journey of a fellowship of characters as they embark on a perilous quest to rid Middle-earth of the dark lord Sauron and his army of orcs.}
\\ \bottomrule
\end{tabular}
}
\vspace{2mm}
\captionof{table}{The challenging examples on LLaVA-1.5, our approach can provide reasonable reponses for the tricky questions which need to fully understand the contexts of the images.}
\label{sup_tab:example1}  
  \end{minipage}
\end{table}

\clearpage
\begin{table}
  \begin{minipage}{0.99\textwidth}
\centering  
\vspace{-4mm}
\scalebox{0.9}{
\begin{tabular}{c p{6.2cm} p{6.2cm} }
\toprule
 \multicolumn{3}{l}{\bf Visual input example, context-related questions:}  \\
\midrule
&  \includegraphics[height=4.5cm,width=6.2cm]{figures/sup_figure2_1.pdf} &  \includegraphics[height=4.5cm,width=6.2cm]{figures/sup_figure2_2.pdf} \\
& Can you explain this meme? & {What kind of school does the boy go to? Answer it and give the rationale.}
\\ \midrule
LLaVA-7B & The meme features a green Kermit the Frog puppet lying on a bed, with a caption that reads, ``Me looks like I gotta wait till 9'' The puppet appears to be relaxing on the bed, possibly waiting for someone or something. The image is a playful and lighthearted representation of a common situation where people might have to wait for a certain time before they can proceed with their plans. & The boy likely goes to a private school, as he is wearing a suit and tie, which is a common attire for students in private schools. The presence of a tie suggests that the school has a formal dress code, and the boy is dressed appropriately for the occasion.
\\ \midrule
\textbf{Ours} & \textcolor{brown}{The image features a green frog, which is a popular character from the television show ``The Muppet Show''}, the frog is sitting on a bed, appearing to be in a relaxed and comfortable position. The puppet is holding its belly and seems to be enjoying its time. The bed has a white sheet, and the scene appears to be set in a cozy bedroom. \textcolor{brown}{The meme is captioned with a humorous message which reads ``Me: I'll do it at 8. Time: 8.05. Looks like I gotta wait till 9 now'', this implies that the puppet is aware of the time and is intentionally delaying any work or task it might be assigned.} The meme is likely meant to be amusing and entertaining, as it combines the idea of a puppet with a human-like sense of humor and awareness of time. & The boy in the image is wearing a tie, which suggests that \textcolor{brown}{he might be attending a school with a formal dress code or a school that requires students to wear ties as part of their uniform. This could be a private school, a religious school, or a school with a strong emphasis on discipline and professionalism.} The presence of a woman, who is wearing a dress, in the image, possibly his mother, indicates that the boy might be going to school with her support and guidance. 
\\ \bottomrule
\end{tabular}
}
\vspace{2mm}
\captionof{table}{The challenging examples on LLaVA-1.5, our approach can provide reasonable reponses for the tricky questions which need to fully understand the contexts of the images.}
\label{sup_tab:example2}  
  \end{minipage}
\end{table}

\clearpage
\begin{table}
  \begin{minipage}{0.99\textwidth}
\centering  
\vspace{-4mm}
\scalebox{0.9}{
\begin{tabular}{c p{6.2cm} p{6.2cm} }
\toprule
 \multicolumn{3}{l}{\bf Visual input example, relationship-aware questions:}  \\
\midrule
&  \includegraphics[height=4.5cm,width=6.2cm]{figures/sup_figure3_1.pdf} &  \includegraphics[height=4.5cm,width=6.2cm]{figures/sup_figure3_2.pdf} \\
& Who is the person on the left? & {Who is the person in the bottom right?}
\\ \midrule
LLaVA-7B & The person on the left is Reese Witherspoon. & The person in the bottom right is a man named Tom Hiddleston.
\\ \midrule
\textbf{Ours} & The person on the left is \textcolor{brown}{Keira Knightley}. & The person in the bottom rights is \textcolor{brown}{Jamie Dornan}.
\\ \bottomrule
\end{tabular}
}
\vspace{2mm}
\captionof{table}{The challenging examples on LLaVA-1.5, our approach can provide reasonable reponses for the tricky questions which need to fully understand the relationships among objects within images.}
\label{sup_tab:example3}  
  \end{minipage}
\end{table}

\clearpage
\begin{table}
  \begin{minipage}{0.99\textwidth}
\centering  
\vspace{-4mm}
\scalebox{0.9}{
\begin{tabular}{c p{6.2cm} p{6.2cm} }
\toprule
 \multicolumn{3}{l}{\bf Visual input example, text-related questions:}  \\
\midrule
&  \includegraphics[height=4.5cm,width=6.2cm]{figures/sup_figure4_1.pdf} &  \includegraphics[height=4.5cm,width=6.2cm]{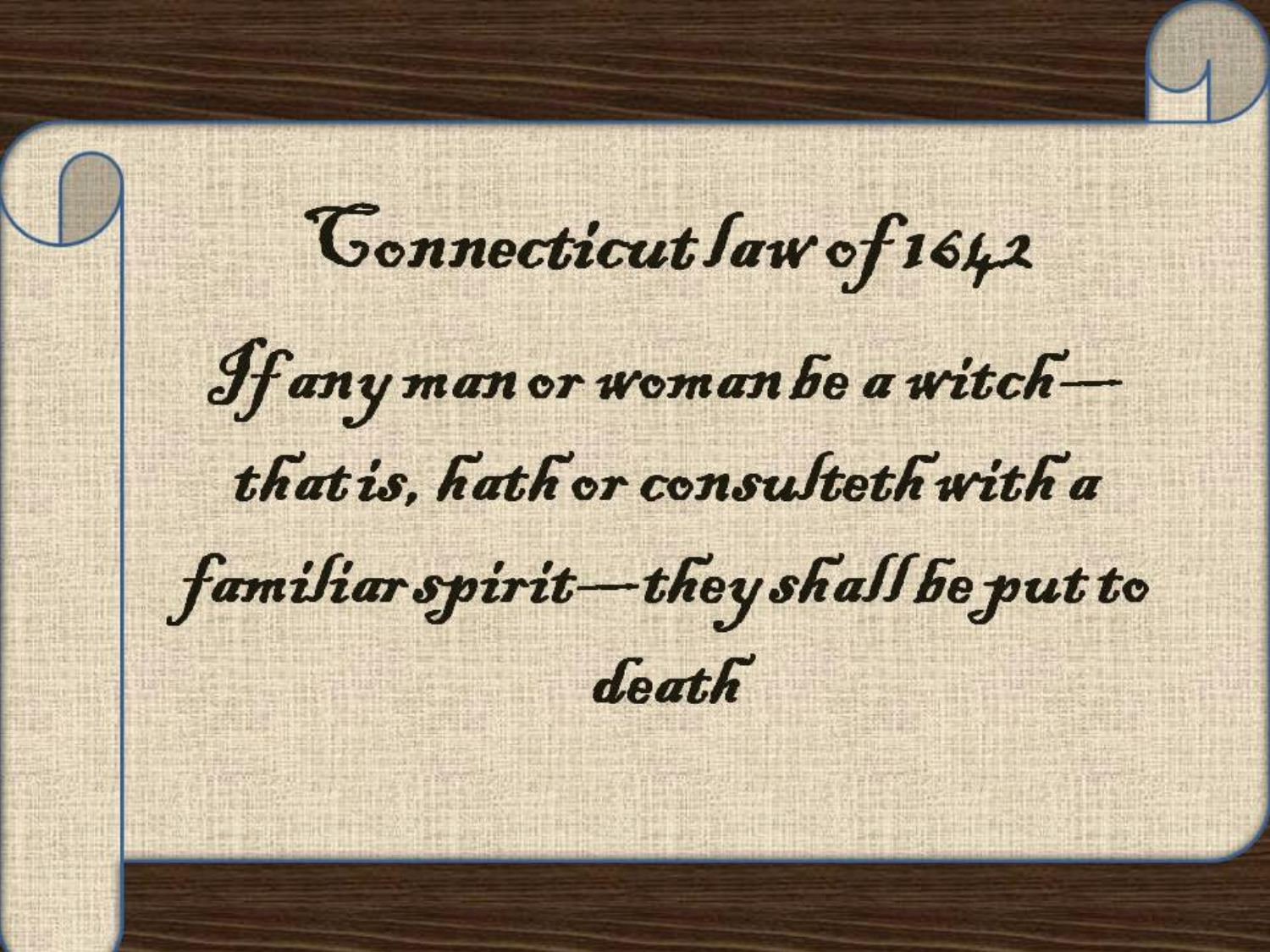} \\
& What are all the scene text in the image? & {What are all the scene text in the image?}
\\ \midrule
LLaVA-7B & The scene text in the image is ``S33020CD''. & The scene text in the image is ``Connecticut Law of 1652''.
\\ \midrule
\textbf{Ours} & The scene text in the image includes the license plate number \textcolor{brown}{``S3302CD''} which is written in white on a black background. & The scene text in the image includes \textcolor{brown}{``Connecticut Law of 1642''}, which states \textcolor{brown}{``If any man or woman be a witch - that is hath or consulteth with a familiar spirit - they shall be put to death''}.
\\ \bottomrule
\end{tabular}
}
\vspace{2mm}
\captionof{table}{The challenging examples on LLaVA-1.5, our approach can provide reasonable reponses for the tricky questions which need to accurately recognize the texts within the images.}
\label{sup_tab:example4}  
  \end{minipage}
\end{table}

\end{document}